\documentclass{article} %
\usepackage[final]{colm2026_conference}

\usepackage{graphicx}
\usepackage{microtype}
\usepackage{hyperref}
\usepackage{url}
\usepackage{booktabs}
\usepackage{wrapfig}
\usepackage{amsmath}
\usepackage{amssymb}
\usepackage{xspace}
\usepackage{pgfplots}
\usepackage{subcaption}
\usepackage{pgfplotstable}
\usepackage{multirow}   %
\usepgfplotslibrary{groupplots}
\pgfplotsset{compat=1.18}
\usetikzlibrary{arrows.meta}
\pgfplotsset{
  lc arch/.style={
    width=0.45\textwidth,
    height=0.38\textwidth,
    grid=major,
    grid style={line width=0.3pt, draw=gray!30},
    axis lines=left,
    tick align=outside,
    tick label style={font=\small},
    label style={font=\small},
    legend style={
      font=\small,
      draw=gray!50,
      fill=white,
      at={(0.03,0.97)},
      anchor=north west,
    },
    mark size=2.5pt,
    line width=0.8pt,
  },
  llama style/.style={
    only marks,
    mark=*,
    color=blue!70!black,
  },
  olmo style/.style={
    only marks,
    mark=square*,
    color=orange!80!red,
  },
}

\usepackage{lineno}
\usepackage{color-edits}
\addauthor{gn}{magenta}

\definecolor{darkblue}{rgb}{0, 0, 0.5}
\hypersetup{colorlinks=true, citecolor=darkblue, linkcolor=darkblue, urlcolor=darkblue}

\title{ Cracks in the Foundation:  Seemingly Minor Architectural \\ Choices Impact Long Context Extension}

\author{
Amanda Bertsch\textsuperscript{1,2} \quad
Luca Soldaini\textsuperscript{1} \quad
Matthew R. Gormley\textsuperscript{2} \quad
Graham Neubig\textsuperscript{2} \\
\textbf{
Hannaneh Hajishirzi\textsuperscript{1,3} \quad
Kyle Lo\textsuperscript{1,3} \quad
Dirk Groeneveld\textsuperscript{1}
}\\
\\
\textsuperscript{1} Ai2 \quad
\textsuperscript{2} Carnegie Mellon University \quad
\textsuperscript{3} University of Washington
}

\newcommand{\poolname}{OlmPool\xspace}
\newcommand{\poolsize}{26\xspace}

\begin{document}

\ifcolmsubmission
\linenumbers
\fi

\maketitle
\begin{abstract}
One might imagine that architectural variations within the dense transformer paradigm have a limited effect on accuracy. However, we demonstrate that this is not the case in the long context setting. Specifically, we show that a set of four minor architectural decisions --- all made by at least one of the Olmo, Llama, and Qwen dense model families --- have a compoundingly negative effect on long context extensibility. Any one of these choices alone has a minor impact on long context performance, but combining three or more can drop the performance downstream by up to 47\%. Furthermore, these differences are not detectable from short-context loss or validation datasets.
We show that much of the variation in long context ability across model families is driven by these architectural features and detectable from applying context extension early in pretraining. 
We demonstrate this with controlled ablations that hold data, tokenizer, and extension recipe fixed while varying normalization, GQA, pretraining context length, and sliding window attention. After over 170,000 GPU hours of training, we release the resulting set of models as \poolname, a set of \poolsize comparable 7B models with checkpoints before and after long-context extension. This pool includes several architectures that outperform the Llama 3 architecture on long context extensibility. In an analysis of our ablation models, we identify patterns in attention sink behavior and attention distributions across context that are attributable to specific architectural differences.

\end{abstract}

\section{Introduction}

Pretraining large language models is an expensive and time-consuming process. From architectural choices to data selection, many design choices carry the potential to shift the behavior of the resulting downstream model --- and these factors can interact in complex ways. 
Because running experiments at full scale is prohibitively expensive, a critical question is how to validate design decisions early in training or at smaller scale. %
This challenge is especially acute for capabilities that are elicited later in the development cycle --- such as reasoning \citep{wang2025octothinkermidtrainingincentivizesreinforcement} or agentic behavior \citep{qin2025aptbenchbenchmarkingagenticpotential} --- since architectural choices must be made long before these capabilities can be directly observed.

Long-context processing is an important instance of this problem. Context length is typically extended by modifying positional embeddings and continuing to pretrain at longer context lengths during a midtraining phase at the end of pretraining \citep{xiong-etal-2024-effective}. Because this phase comes late in the development cycle, practitioners must commit to architectural decisions before they can observe how those decisions affect long-context behavior. Compounding this, most long-context extension recipes are developed on a small set of base models: the majority of works focus on extending Llama family models (\citet{fu2024dataengineeringscalinglanguage, gao2025trainlongcontextlanguagemodels, lu2024controlledstudylongcontext, chen2024longloraefficientfinetuninglongcontext, peng2026yarnefficientcontextwindow}, \textit{inter alia}), with relatively few considering other architectures (e.g. \citet{ding2024longropeextendingllmcontext, zhao2024longskyworktrainingrecipeefficiently, hu2024longreciperecipeefficientlong}). As a result, it is unclear how broadly existing recipes transfer, or whether the base architecture itself is a decisive factor in downstream long-context performance, even when comparing only transformer models.\footnote{In parallel, recent work on architectures outside of the standard transformer has been motivated by improving long context performance (e.g \citet{gu2024mambalineartimesequencemodeling, yang2025gateddeltanetworksimproving, peng2023rwkvreinventingrnnstransformer}); we focus here on variations \textit{within} the transformer family.} %

\begin{figure*}[t]
\centering
\begin{tikzpicture}
\begin{axis}[
  width=\textwidth,
  height=0.4\textwidth,
  ybar, bar width=6pt,
  ylabel={HELMET 32K},
  xlabel={Models sorted by HELMET 32K},
  xtick=\empty,
  xmin=-1, xmax=26,
  ymin=0, ymax=65,
  axis on top=false,
  legend style={at={(0.5,1.02)}, anchor=south, legend columns=-1, font=\small},
  legend cell align=left,
]

  \addplot[fill=gray!60, draw=none, bar shift=0pt, forget plot]
    table[col sep=comma, x=idx, y=val]{figures/data/helmet_feat0.csv};
  \addlegendimage{fill=gray!60, draw=none, area legend}\addlegendentry{0 features}

  \addplot[fill=blue!40!white, draw=none, bar shift=0pt, forget plot]
    table[col sep=comma, x=idx, y=val]{figures/data/helmet_feat1.csv};
  \addlegendimage{fill=blue!40!white, draw=none, area legend}\addlegendentry{1 feature}

  \addplot[fill=teal!70, draw=none, bar shift=0pt, forget plot]
    table[col sep=comma, x=idx, y=val]{figures/data/helmet_feat2.csv};
  \addlegendimage{fill=teal!70, draw=none, area legend}\addlegendentry{2 features}

  \addplot[fill=orange!80!black, draw=none, bar shift=0pt, forget plot]
    table[col sep=comma, x=idx, y=val]{figures/data/helmet_feat3.csv};
  \addlegendimage{fill=orange!80!black, draw=none, area legend}\addlegendentry{3 features}

  \addplot[fill=red!80!black, draw=none, bar shift=0pt, forget plot]
    table[col sep=comma, x=idx, y=val]{figures/data/helmet_feat4.csv};
  \addlegendimage{fill=red!80!black, draw=none, area legend}\addlegendentry{4 features}

  \node[white, font=\bfseries\tiny] at (axis cs:16, 2.5) {O};
  \node[white, font=\bfseries\tiny] at (axis cs:22, 2.5) {L};
  \node[white, font=\bfseries\tiny] at (axis cs:13, 2.5) {Q};

  \draw[<->, thick] (axis cs:-0.0, 29.87) -- (axis cs:-0.0, 56.38);
  \node[right, font=\small] at (axis cs:-0.0, 43.1) {$\Delta$26.5};

\end{axis}
\end{tikzpicture}
\caption{HELMET 32K scores across all \poolname models with identical data and context extension strategy, sorted worst to best. The colors indicate the count of (individually minor) choices made that downweight long context performance; the combination of these features can dramatically degrade performance, up to 26.5 points on HELMET. (Q), (O), and (L) indicate the Qwen 3, Olmo 3, and Llama 3 architectures; none of these is optimal.}
\label{fig:helmet_feature_count}
\end{figure*}
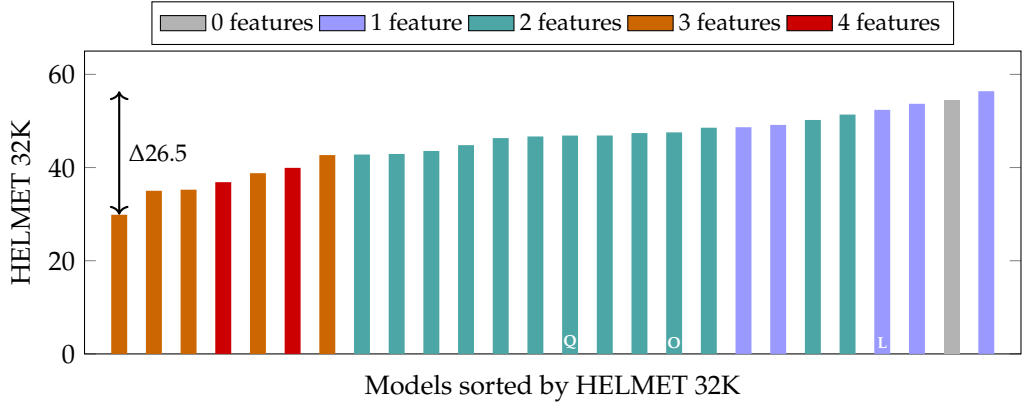

In this work, we demonstrate that a small set of cross-model-family architectural variations account for substantial variation in downstream long context performance by performing a set of data- and optimization-controlled pretraining experiments.  We first show that short context performance is not sufficient to predict long context performance by training a sweep of models with the same short context behavior but dramatically variable long context behavior; 
then, we study how normalization decisions, the use of GQA, sliding window attention, and the pretraining context length shift long context performance.  
We select only values for these four factors that have been used in Llama 2, Llama 3, Qwen 3, or Olmo 3, and pretrain a sweep of \poolsize models, \poolname, which represent varying choices for each factor. We show that even interpolating in this narrow design space can result in dramatically divergent long context performance (e.g. in Figure~\ref{fig:helmet_feature_count}). 
This presents a tradeoff: each of these design decisions has desirable properties for training efficiency, inference cost, or stability, while also negatively impacting long context extensibility.
We identify minimal pairs of architectural changes that substantially damage downstream long context performance, characterize the compounding impact of applying multiple of these architectural changes at once, and analyze the attention patterns of these model pairs.

\section{Setting}
We perform a set of controlled pretraining experiments to construct \poolname,\footnote{An \textit{olm} is a type of cave-dwelling aquatic salamander.}
 a set of \poolsize comparable models in the 7-8B parameter range. Appendix~\ref{appendix:pool_desc} provides full descriptions. %

\subsection{Architectural choices ablated} 

We consider four primary architectural design decisions: normalization strategy, grouped-query attention, the use of sliding windows, and pretraining context length. These features were selected because they differ across several major recent model releases and have explicit connections to the attention mechanism or context length.

\paragraph{Normalization} The two normalization factors we consider are layernorm ordering and the presence of QK norm. The observation that QK norm can limit long context performance was first made by \citet{yang2025rope}; we replicate this finding in our setting and further explore how specific variants of QK norm and norm order play a role. 
 
QK norm is generally added to improve training stability (e.g. in \citet{chameleonteam2025chameleonmixedmodalearlyfusionfoundation, marin2025retro32b}) and is often implemented layerwise, as a normalization step applied before the concatenated query matrix is split for specific attention heads. This may be implemented as an RMS norm:

\begin{equation}
    \hat{Q} = \frac{Q}{\text{RMS}(Q)}\,\gamma^Q, \qquad \hat{K} = \frac{K}{\text{RMS}(K)}\,\gamma^K
\end{equation}

Where $\gamma^Q, \gamma^K$ are normalization parameters learned for each layer. 
This \textit{layerwise} implementation of QK norm is used in Olmo 2 and 3 \citep{olmo20252olmo2furious, olmo3_2025}. We also consider the headwise variant of QK norm, used by Qwen 3 \citep{yang2025qwen3technicalreport}, Gemma 3 \citep{gemmateam2025gemma3technicalreport}, and Marin 32B \citep{marin2025retro32b}. Headwise QK norm applies normalization separately to each attention head's queries and keys, learning per-head values $\gamma_h^Q, \gamma_h^K$ for $h = 1,\dots,H$:

\begin{equation}
    \hat{Q}_h = \frac{Q_h}{\mathrm{RMS}(Q_h)}\,\gamma^Q_h, \qquad
    \hat{K}_h = \frac{K_h}{\mathrm{RMS}(K_h)}\,\gamma^K_h
\end{equation}

A related question is whether to place the layernorm before or after the sublayer (i.e. prenorm or post-sublayer-norm\footnote{Closely related to perinorm, which normalizes a sublayer's inputs \textit{and} outputs \citep{kim2025perilnrevisitingnormalizationlayer}.}). Applying post-sublayer-norm without QK norm can lead to training divergence because models that apply normalization after the sublayer are more sensitive to gradient instability caused by large or high-variance attention logits.%
\footnote{We confirm this experimentally as well; in runs with post-sublayer-norm and no QK norm, training consistently diverged before 140B tokens.}

\paragraph{Grouped query attention (GQA).} GQA \citep{ainslie2023gqatraininggeneralizedmultiquery} increases inference efficiency by reusing the same key-value matrices for multiple query heads in the same layer, reducing the size of the key-value cache. Typical GQA models share 8 key-value heads for 32 query heads \citep{grattafiori2024llama3herdmodels, yang2025qwen3technicalreport}. Note that because GQA shares some $W_K$ and $W_V$ parameters, it reduces the total capacity of the network; when this occurs, we adjust the intermediate size slightly to result in the same total parameter count. This should benefit models with GQA in our comparisons. %

\paragraph{Sliding window attention (SWA).} Modern sliding window attention implementations generally intersperse layers of local window attention with layers of full attention in a many-to-one pattern  \citep{olmo3_2025, gemmateam2025gemma3technicalreport}, which reduces the KV cache size at inference time. We use the Olmo 3 configuration, which has 3 local attention layers of 4096 context for every 1 global attention layer.

\paragraph{Pretrained context length.} \citet{zhao2024longskyworktrainingrecipeefficiently} observe that long context ability is impacted by the pretraining context length, with models trained at a longer context length better able to adapt to context extension. %
Conversely, shorter context lengths generally allow for faster training throughput. We pretrain Olmo and Llama variants at 4096 context length, matching the context length of the prior generation for each model. This allows us to compare models trained at 4K directly with models that use a 4K sliding window on 8K context.%

\subsection{Training}

We pretrain each model for 140B tokens (the Chinchilla-optimal amount of data \citep{hoffmann2022trainingcomputeoptimallargelanguage}). 
Then, we adjust the RoPE theta for context extension \citep{xiong2023effectivelongcontextscalingfoundation} and continue pretraining for 10B tokens on 64K context data from the Longmino mix \citep{olmo3_2025}, annealing the learning rate to 0. We hold the data selection and ordering, the learning rate and schedule, and the tokenizer constant across all models in \poolname. We reuse the same optimization learning rate and schedule as Olmo 3 \citep{olmo3_2025} and further discuss the rationale for standardizing optimization in Appendix~\ref{appendix:training_details}.

Ideally, we would also standardize initialization. However, because not all models have the same parameters (e.g. if GQA or QK norm are added), we cannot exactly synchronize the initializations. Where possible, we reuse the same initalization across runs; if there is a minor difference in parameterization, we use the remainder of the same initialization and only re-initialize the new parameters. Appendix~\ref{appendix:pool_desc} identifies the initialization for each model, and we further discuss the impact of model initialization in Appendix~\ref{appendix:additional_features}.

 \subsection{Evaluation}
 We evaluate all models downstream on 3 popular measures of long context performance: RULER \citep{hsieh2024ruler}, which represents a sweep of synthetic Needle-in-a-Haystack (NIAH) style tasks of increasing complexity; HELMET \citep{yen2025helmet}, which additionally considers in-context learning, reranking, and question-answering tasks; and LongPPL \citep{fang2025wrong}, a variant of perplexity that only considers tokens that require long-range dependencies to predict. Because we are working with weak models early in pretraining, we consider only the subtasks from HELMET that do not require generating long spans of text to evaluate with an LM judge. We observe that all three measures correlate closely; for readability, we primarily report HELMET at 32K in the main text, unless trends differ across the three measures. %
 To compare model architectures, we construct multiple paired comparisons, holding constant all other factors, wherever possible. We also fit linear regressions from architectural features, short-context metrics, and observed attention distributions to measure the predictive power of each feature across \poolname. %

\section{Short context metrics are not predictive of long context performance}
\begin{figure*}
    \centering
    \begin{subfigure}[t]{0.48\textwidth}
            \begin{tikzpicture}
\begin{axis}[
    width=\textwidth,
    height=0.8\textwidth,
    xlabel={Short context evals (BPB) ($\downarrow$)},
    ylabel={HELMET 32K},
    xmin=0.661, xmax=0.702,
    tick align=outside,
    tick pos=left,
    grid=major,
    grid style={dashed, gray!30},
]

  \addplot[only marks, mark=*, mark size=2.5pt,     color=blue!70!black]
    table[col sep=comma, x=SC_avg, y=HELMET_32K]
    {figures/data/plot1_all.csv};

\end{axis}
\end{tikzpicture}  %

    \end{subfigure}
    \begin{subfigure}[t]{0.48\textwidth}
            \begin{tikzpicture}
\begin{axis}[
    width=\textwidth,
    height=0.8\textwidth,
    xlabel={Pre-extension HELMET 8K ($\uparrow$)},
    ylabel={HELMET 32K ($\uparrow$)},
    xmin=8,  xmax=54,
    ymin=25, ymax=62,
    xtick={10,20,30,40,50},
    ytick={30,40,50,60},
    tick align=outside,
    tick pos=left,
    grid=major,
    grid style={dashed, gray!30},
]

\addplot[
    only marks,
    mark=*,
    mark size=2.5pt,
    color=blue!70!black,
] coordinates {
    (17.70, 47.55)
    (44.85, 48.66)
    (31.22, 29.87)
    (18.84, 42.68)
    (44.28, 52.38)
    (49.50, 56.38)
    (46.39, 49.14)
    (48.80, 54.42)
    (39.09, 42.80)
    (17.98, 46.67)
    (31.74, 38.79)
    (44.76, 46.88)
    (46.06, 48.55)
    (43.53, 53.69)
    (38.43, 46.32)
    (37.66, 51.37)
    (44.39, 44.81)
    (39.39, 43.56)
    (28.12, 35.25)
    (14.25, 47.40)
    (12.99, 42.91)
    (12.99, 35.01)
    (12.57, 39.93)
    (16.16, 36.86)
    (49.10, 46.86)
};

\end{axis}
\end{tikzpicture}
            \label{subfig:helmet_sc}
            
    \end{subfigure}
   \caption{Benchmark scores pre-extension largely fail to predict long-context benchmark scores post-extension, even when evaluating on the shorter-context version of the same benchmark. Each point is a model from \poolname. %
   }
    \label{fig:evals_vs_lc}

\end{figure*}
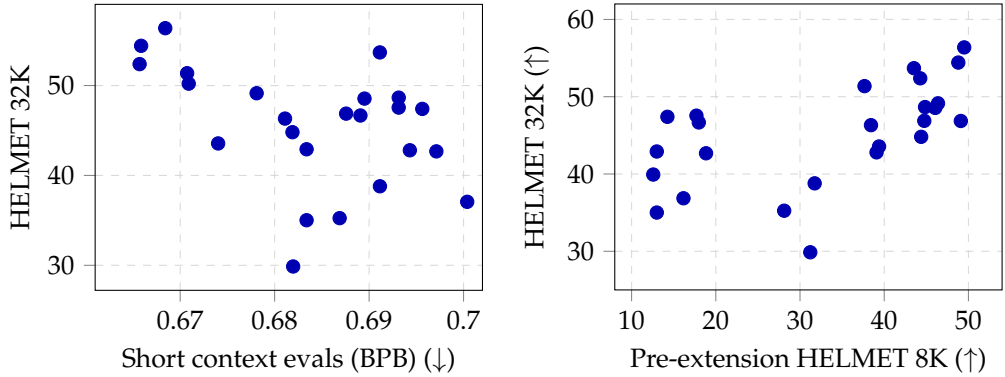

 The models in \poolname demonstrate that seemingly small changes in the model recipe have a dramatic downstream effect on long context extensibility. The performance of these models ranges from 29.9 to 56.4 on HELMET at 32K, or 44.7 to 67.7 on RULER at 32K. %
 
 Can this downstream effect be predicted from pretraining metrics? We observe that short context metrics are surprisingly poor predictors of long-context behavior. %
 We consider a sweep of measures to try to predict long context performance downstream, demonstrating that standard pretraining metrics are insufficient to predict long context performance. 
\paragraph{Intrinsic metrics.} We measure the loss for each training run at the end of pretraining and at the end of long context extension and compute the correlation between these values and the resulting HELMET score. 
Training loss correlates only weakly with downstream long context performance ($R^2=0.29)$; surprisingly, the loss during pretraining is \textit{more predictive} of downstream score than the loss during context extension ($R^2=0.06$).  %
We measure perplexity of the pre-context-extension model over 11 held-out text samples from differing domains \citep{magnusson2024paloma}; these scores range from completely uncorrelated to weakly correlated with downstream long context scores. The best-correlated splits do not align with common understanding of the types of data that require long-range dependencies: a sample of data from WikiText \citep{merity2016pointer} is the most correlated ($R^2=0.39$), with perplexity of samples from academic texts and code showing little to no correlation with long context behavior downstream ($0.01 \le R^2 \le 0.20$).\footnote{These trends hold even if we measure correlation between perplexity and LongPPL, which is a more intrinsic metric of long context quality. See a full per-dataset breakdown in Appendix~\ref{appendix:short_context_metrics}.}

\paragraph{Downstream evaluations.} During pretraining, models are often periodically evaluated on a set of development benchmarks. We evaluate on a set of 16 in-loop benchmarks, detailed in Appendix~\ref{appendix:short_context_metrics}; because early pretraining checkpoints are often inconsistent at answering in multiple-choice format \citep{bunn-etal-2025-fine}, we score by the bits-per-byte on the correct answer for each benchmark question instead of accuracy \citep{heineman2025signal}. Figure~\ref{fig:evals_vs_lc} (left) shows the average of these evaluations graphed against downstream HELMET score; there is a slight correlation between the two ($R^2=0.17$), but this fails to exceed the predictiveness of the best perplexity measures. %
Note that all scores are very close together: in our setting, where models are trained with the same data and very similar architectures, little difference is observable between models pre-context-extension. 
 Clearly, standard in-loop evaluations are not sufficient to provide signal for downstream long context extensibility. 

Is this merely an issue of choosing the wrong benchmarks?  We consider
a benchmark more directly related to our evaluations for performance downstream: the shortest context split of HELMET, which, at 8K, is possible for most of our models to process without context extension. Recall that these models are being evaluated pre-anneal and early in training, so we expect scores to be quite poor. %
Figure~\ref{fig:evals_vs_lc} (right) graphs these pre-extension scores against the post-extension performance; while HELMET scores pre-extension are both more variable and slightly more predictive ($R^2=0.32$) than other short-context evaluations, they still fail to predict double-digit swings in HELMET performance downstream. %

\section{Impact of architectural choices}
\label{sec:main_results}

Clearly, the differing performance of these models is not solely attributable to how well each model has fit the pretraining corpus. We consider the individual effect of each of the four factors we have identified in turn and show that their behavior is best modeled as an additive impairment to long context: the individual features selected do not matter nearly as much as the number of long-context-inhibiting features present.
In Appendix~\ref{appendix:additional_features}, we also discuss the impact of additional features that we considered: pretraining with linear layers quantized to float8 and changing the pretraining run's random initialization.

\begin{figure*}[t]
\centering
\begin{tikzpicture}
\begin{groupplot}[
    group style={
        group size=3 by 1,
        horizontal sep=1.8cm,
    },
    width=4.5cm,
    height=5cm,
    xlabel={\# KV Heads},
    xtick={4,8,16,32},
    xticklabels={4,8,16,32},
    xmin=2, xmax=36,
    tick align=outside,
    tick pos=left,
    grid=major,
    grid style={dashed, gray!30},
    every axis plot/.style={thick, mark=*, mark options={fill=white, line width=1.5pt}},
]

\nextgroupplot[
    ylabel={HELMET 32K ($\uparrow$)},
    ymin=48, ymax=58,
]
\addplot[color=blue!70] coordinates {
    (4,  49.14)
    (8,  52.38)
    (16, 56.38)
    (32, 54.42)
};

\nextgroupplot[
    ylabel={RULER 32K ($\uparrow$)},
    ymin=60, ymax=69,
]
\addplot[color=red!70] coordinates {
    (4,  61.21)
    (8,  64.58)
    (16, 67.72)
    (32, 67.51)
};

\nextgroupplot[
    ylabel={LongPPL ($\downarrow$)},
    ymin=3.38, ymax=3.65,
]
\addplot[color=green!60!black] coordinates {
    (4,  3.6318)
    (8,  3.5659)
    (16, 3.5020)
    (32, 3.4100)
};

\end{groupplot}
\end{tikzpicture}
\caption{ GQA is harmful to long context performance in our setting, even when adjusting for comparable parameter count. Variants of Llama that increase the number of KV heads improve performance on long context benchmarks. 8 KV heads is the Llama GQA configuration, and 32 KV heads indicates no GQA. Note that the model with 16 KV heads is slightly larger than the other models in this comparison (8.1B vs 7.7-7.8B).}
\label{fig:llama_gqa}
\end{figure*}
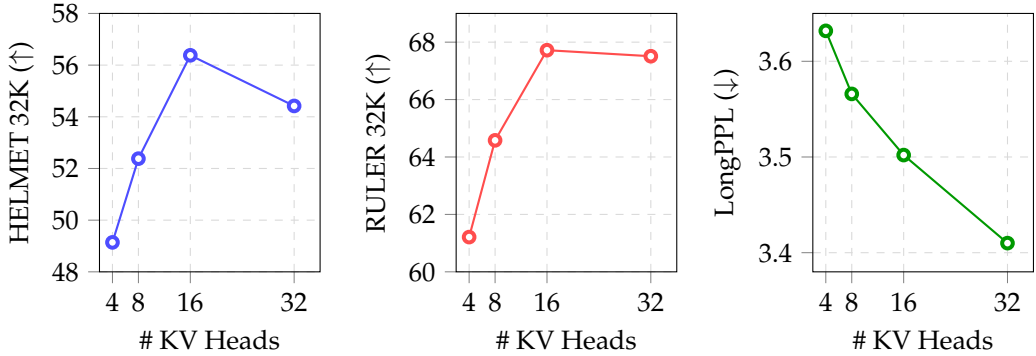

 \paragraph{All four features reduce long context capability downstream.} In paired comparison runs, each architectural feature results in some degradation of long context performance downstream.
Pretraining at 4096 instead of 8192 context length and training with sliding window layers result in modest average degradations of 1-2 points on HELMET at 32K. 
Figure~\ref{fig:llama_gqa} shows that pretraining with GQA configurations that use increasingly aggressive degrees of query sharing (i.e. decreasing the number of KV heads) degrades performance from the Llama 3 configuration, and pretraining with \textit{more} KV heads than the Llama 3 architecture improves performance. 
But the largest individual performance impact by far arises from normalization choice.
On the Olmo architecture, changing Olmo 3's choice of QK norm and post-sublayer-norm to prenorm results in a 6 point gain on HELMET; conversely, adding these features to Llama 3's architecture results in a 3.8 point drop.\footnote{We find that the choice of headwise versus layerwise QK norm and the ordering of prenorm vs post-sublayer-norm have some effect as well, although the decision to apply QK norm at all dominates; for more details, see Appendix~\ref{appendix:additional_features}.}

\paragraph{Minor effects of architecture compound.} With the exception of QK norm, most individual architectural features have relatively minor effects when they are the only potentially detrimental feature tested. However, when paired together, these features can have a much more significant impact.
For instance, adding a sliding window has a minor negative effect on a model without GQA (-1.1 points from the full attention configuration). However, in multiple comparison runs, adding sliding window layers to a model with GQA results in a dramatic drop in performance: -9 points on average. The worst-scoring runs in \poolname combine two or more features that limit the expressivity of attention ---  e.g. the single worst configuration combines GQA, sliding windows, and headwise QK norm, for a total effect far worse than the sum of these individual impacts. 

We find that downstream long context performance can be well-estimated by simply counting how many of these four architectural choices are present in a \poolname model --- this single numerical feature is the single most predictive feature for downstream long context performance (in-sample $R^2=0.67$, LOO $R^2=0.61$), outperforming even a linear regression over the four individual axes of architectural variation. It is not any single feature which results in catastrophically poor long context extensibility,  but the combination of several features that each reduce the expressivity of attention.

Figure~\ref{fig:helmet_feature_count} also shows that combining all four features is not necessarily worse than combining three features in our setting. We hypothesize that this may be because at least some of these features impact long context abilities in overlapping ways: for instance, sliding window attention restricts the model to only 4K context at some layers during pretraining, and pretraining at 4K restricts the model to only 4K context at \textit{all} layers during pretraining. However, the predictor that counts the presence of any of the four features remains more predictive than any predictor that collapses two of the features into the same category.

 \paragraph{Llama 3 is a particularly good architecture for long context.} While it was previously unclear whether the ease of extensibility for Llama 3 was due to architectural or data factors (since the pretraining data for Llama 3 was not publicly disclosed), %
  we find evidence that this is primarily an architectural phenomenon. %
  The Llama 3 architecture model is one of the best models in the design space. 
  This suggests that context extension recipes developed with this model may require additional effort to apply to other architectures --- for instance, our results show that same context extension recipe applied to the Llama 3, Qwen 3, and Olmo 3 architectures is far more effective for the Llama-like model, even when the other models were pretrained for the same duration and data as the Llama-like model. This also helps explain our empirical observation that Olmo 3 Base is more challenging to context extend than Llama 3 Base.%

\section{Analysis}

\subsection{Is this measuring token efficiency or a fundamental capability gap?}

Our main experiments compare models after the same 10B token context extension. However, it's possible that extending with a much longer context extension phase would wash out these differences. %
To test this, we choose three representative models at different quality points in \poolname: the models with the architectures of Llama 3, Olmo 3, and the worst-scoring architecture on HELMET downstream. On this trio of models, we conduct 1B and 50B context extensions and compare performance with increasing amounts of data.\footnote{Note that because we perform each context extension run as an anneal to a learning rate of 0, the 1B, 10B, and 50B extensions are separate runs, not points sampled from one long extension.} 
Figure~\ref{fig:token_efficiency} shows the performance of these models with increasing amounts of data. As expected, all models see better long context performance with 50B token extensions; however, this effect is not enough to compensate for architecture-driven differences in base quality. Even after 50B tokens, the worst architecture does not reach the same performance as the Llama architecture achieves after 1B tokens, and the differences between the three architectures remains relatively stable.
While it's possible that context extensions at a more extreme scale could overpower these architectural effects, we see no evidence of this occurring up to the 50B token scale, where the context extension phrase represents 26\% of the total tokens seen by the model. %

\begin{figure*}[t]
    \centering
    \begin{minipage}{0.48\linewidth}
        \centering
        \begin{tikzpicture}
\begin{axis}[
  width=\textwidth,
  height=0.8\textwidth,
  grid=major,
  grid style={line width=0.3pt, draw=gray!25},
  axis lines=left,
  tick align=outside,
  tick label style={font=\small},
  label style={font=\small},
  xmode=log,
  log basis x=10,
  xtick={1,10,50},
  xticklabels={1B,10B,50B},
  xlabel={Context extension tokens},
  ylabel={HELMET score},
  ymin=0, ymax=65,
  xmin=0.7, xmax=70,
  mark size=2.5pt,
  line width=1pt,
  legend style={font=\small, at={(0.5,1.02)}, anchor=south, legend columns=-1},
]

\addplot[mark=*, color=blue!70!black]
  coordinates {(1,30.45) (10,29.87) (50,39.02)};
\addlegendentry{Worst}

\addplot[mark=square*, color=orange!85!red]
  coordinates {(1,36.69) (10,47.55) (50,49.87)};
\addlegendentry{Olmo}

\addplot[mark=triangle*, color=green!55!black]
  coordinates {(1,48.42) (10,52.38) (50,56.64)};
\addlegendentry{Llama}

\end{axis}
\end{tikzpicture}
        \caption{Performance on HELMET after 1B, 10B, or 50B token extension for three representative runs (the worst architecture, the Olmo 3 architecture, and the Llama 3 architecture). Longer context extension fails to wash out architectural differences.}
        \label{fig:token_efficiency}
    \end{minipage}
    \hfill
    \begin{minipage}{0.48\linewidth}
        \centering
        \begin{tikzpicture}
\begin{axis}[
  width=\textwidth,
  height=0.8\textwidth,
  grid=major,
  grid style={line width=0.3pt, draw=gray!25},
  axis lines=left,
  tick align=outside,
  tick label style={font=\small},
  label style={font=\small},
  xmode=log,
  log basis x=10,
  xtick={70,140,280,2000},
  xticklabels={70B,140B,280B,2T},
  xlabel={Pretraining tokens},
  ylabel={HELMET score},
  ymin=0, ymax=70,
  xmin=50, xmax=2500,
  mark size=2.5pt,
  line width=1pt,
  legend style={font=\small, at={(0.5,1.02)}, anchor=south, legend columns=-1},
]

\addplot[mark=square*, color=orange!85!red]
  coordinates {(70,38.64) (140,47.55) (280,50.91) (2000,61.43)};
\addlegendentry{Olmo-like}

\addplot[mark=*, color=blue!70!black]
  coordinates {(70,28.06) (140,29.87) (280,34.18) (2000,45.09)};
\addlegendentry{Worst}

\end{axis}
\end{tikzpicture}
        \caption{HELMET score at extensions performed after progressively more pretraining tokens for two training runs. The difference in long context behavior is consistent, especially from 140B onwards.}
        \label{fig:early_pretraining}
    \end{minipage}
\end{figure*}
\subsection{Do these differences persist in longer pretraining runs?}

To be able to measure many architectural configurations, we extend models after only 140B tokens of pretraining. However, most modern models are trained far longer, with long context extension often applied after the model has seen trillions of tokens \citep{olmo3_2025}. Would these effects wash out with a longer pretraining run?

We measure the difference between two configurations of interest on much longer pretraining runs.
At 70B, 140B, 280B, and 2T tokens into a longer pretraining run, we adjust the RoPE theta and perform a 10B anneal on long context (64K) data.
Figure~\ref{fig:early_pretraining} shows the relative behavior of these models on downstream long context evaluations at each point. We show that nontrivial long context performance can be recovered from extensions at least as early as 70B, and the relative performance of the two architectures remains fairly consistent over the course of pretraining. Because we observe that the two models are closer in performance at 70B than at any point 140B or later, we train all other \poolname models to 140B tokens before extending to ensure we are not missing differences across models.

\subsection{Do these differences persist across context extension strategies?}
\begin{wrapfigure}{r}{0.48\textwidth}
  \centering
  \centering
  \colorlet{olmocol}{orange!85!red}
  \colorlet{llamacol}{green!55!black}
  \colorlet{bestcol}{violet!85!black}
  \pgfplotstableread[col sep=comma]{figures/data/extension_comparison.csv}\helmetdata
  \begin{tikzpicture}
  \begin{axis}[
      width=\linewidth, height=5cm,
      ybar,
      bar width=7pt,
      symbolic x coords={8K,16K,32K,64K},
      xtick=data,
      enlarge x limits=0.18,
      ymin=0, ymax=63,
      ytick distance=20,
      clip=false,
      ylabel={HELMET score},
      axis x line*=bottom,
      axis y line*=left,
      axis line style={-{Stealth[length=7pt]}},
      x axis line style={-},   %
      ymajorgrids=true,
      grid style={gray!25, line width=0.4pt},
      tick align=outside,
      tick style={black},
      legend style={
        at={(0.5,1.03)}, anchor=south,
        legend columns=-1,
        draw=black, line width=0.6pt, fill=white,
        inner xsep=6pt, inner ysep=3pt,
        font=\small,
        /tikz/every even column/.append style={column sep=8pt},
      },
  ]
  \addplot[draw=none, fill=olmocol!30,  bar shift=-9pt, forget plot] table[x=context, y=olmo_base]  {\helmetdata};
  \addplot[draw=none, fill=olmocol,     bar shift=-9pt, forget plot] table[x=context, y=olmo_yarn]  {\helmetdata};
  \addplot[draw=none, fill=llamacol!30, bar shift=0pt,  forget plot] table[x=context, y=llama_base] {\helmetdata};
  \addplot[draw=none, fill=llamacol,    bar shift=0pt,  forget plot] table[x=context, y=llama_yarn] {\helmetdata};
  \addplot[draw=none, fill=bestcol!30,  bar shift=9pt,  forget plot] table[x=context, y=gqa_base]   {\helmetdata};
  \addplot[draw=none, fill=bestcol,     bar shift=9pt,  forget plot] table[x=context, y=gqa_yarn]   {\helmetdata};
  \addlegendimage{area legend, draw=none, fill=olmocol}
  \addlegendentry{Olmo}
  \addlegendimage{area legend, draw=none, fill=llamacol}
  \addlegendentry{Llama}
  \addlegendimage{area legend, draw=none, fill=bestcol}
  \addlegendentry{Best}
  \end{axis}
  \end{tikzpicture}
  \caption{Solid color indicates the YaRN 2-stage extension performance; shaded is the NTK 1-stage extension performance. %
  }
  \label{fig:yarn}
\end{wrapfigure}
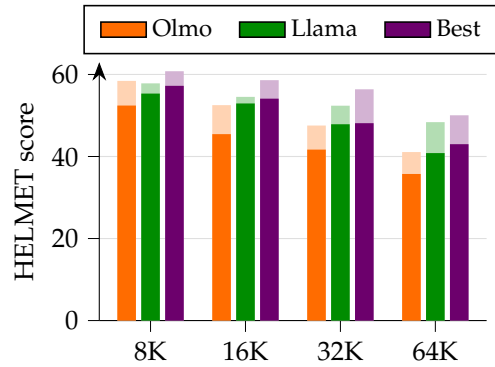
To understand if the differences we observe are sensitive to context extension strategy, we consider an alternate long context extension recipe. 
We perform a two-stage context extension using YaRN \citep{peng2026yarnefficientcontextwindow}.%
We perform continued pretraining at 32K for 5B tokens and then at 64K for 5B tokens, over the course of a combined linear anneal of the learning rate to 0. We warm up the learning rate for 200 steps at the start of the first phase, and use the same YaRN factor across phases.

We perform this modified extension on 3 models: the Olmo-like model, the Llama-like model, and the best model in \poolname (measured by HELMET at 32K after our standard extension). Figure~\ref{fig:yarn} compares these extensions with the extension to 64K used in the primary \poolname analysis. 
This strategy uniformly underperforms our other context extensions, likely because of the reduced number of tokens at 64K context. However, the same relative ranking of architectures is recovered by both methods, and the gap between architectures is larger with the \textit{better} context extension method.

\subsection{Attention behaviors in \poolname}
Armed with a set of models that differ in downstream performance, we now seek to understand the finegrained differences between these models. We compute a number of statistics of the attention distribution across all models, measured over a set 
of 100 long documents selected at random from an even split of Project Gutenberg books, government reports \citep{huang2021efficientattentionslongdocument}, FineWeb PDFs \citep{penedo2024finewebdatasetsdecantingweb}, and legal texts \citep{hendersonkrass2022pileoflaw}. We measure the entropy of the attention distribution at each attention head at positions 1K, 4K, 16K, and 32K. %
We also measure the percentage of attention mass in the attention sink (defined as the first 100 tokens of visible context) and the local context (defined as the 100 tokens immediately preceding the current token).

\begin{figure*}[t]
\centering
\begin{tikzpicture}
\begin{groupplot}[
    group style={
        group size=2 by 1,
        horizontal sep=2cm,
    },
    width=7cm,
    height=6.5cm,
    xlabel={Attention Entropy @ 32K},
    ylabel={\% Sink Attention},
    tick align=outside,
    tick pos=left,
    grid=major,
    grid style={dashed, gray!30},
    ymin=0, ymax=23,
    ytick={0,5,10,15,20},
]

\nextgroupplot[
    title={\textbf{Full-Attention Layers}},
    xmin=4.9, xmax=8.5,
    xtick={5,6,7,8},
    legend style={
        at={(1.05,-0.40)},
        anchor=north,
        legend columns=3,
        column sep=0.6cm,
        draw=none,
        font=\small,
    },
]

\addplot[only marks, mark=*, mark size=2pt, color=blue!70] coordinates {
    (5.156, 20.680)
    (5.261, 19.251)
    (5.183, 19.641)
    (5.338, 18.923)
    (5.363, 11.008)
    (5.554, 12.440)
    (5.453, 11.441)
};
\addlegendentry{no QKNorm}

\addplot[only marks, mark=square*, mark size=2pt, color=red!65] coordinates {
    (7.664,  4.093)
    (7.541,  5.024)
    (7.784,  4.127)
    (7.658,  4.472)
    (7.612,  4.256)
    (7.701,  3.939)
    (7.730,  4.025)
    (6.927,  4.961)
    (6.954,  7.081)
    (7.691,  3.645)
    (7.682,  4.972)
    (7.688,  4.781)
    (7.911,  4.217)
    (7.927,  3.732)
    (8.185,  3.546)
};
\addlegendentry{QKNorm}

\addplot[only marks, mark=triangle*, mark size=2.5pt, color=orange!80!red] coordinates {
    (6.648,  5.440)
    (6.033,  5.234)
    (5.524, 10.706)
};
\addlegendentry{headwise QKNorm}

\nextgroupplot[
    title={\textbf{SWA Layers}},
    ylabel={},
    xmin=4.4, xmax=6.5,
    xtick={4.5,5.0,5.5,6.0,6.5},
]

\addplot[only marks, mark=*, mark size=2pt, color=blue!70] coordinates {
    (4.542, 11.008)
    (4.760, 12.440)
    (4.749, 11.441)
};

\addplot[only marks, mark=square*, mark size=2pt, color=red!65] coordinates {
    (5.899,  4.093)
    (6.173,  4.127)
    (5.887,  4.472)
    (5.905,  4.256)
    (6.184,  3.939)
    (5.326,  4.961)
    (6.204,  3.645)
    (5.823,  4.217)
    (6.103,  3.732)
    (6.089,  3.546)
};

\addplot[only marks, mark=triangle*, mark size=2.5pt, color=orange!80!red] coordinates {
    (5.039,  5.440)
    (4.745,  5.234)
};

\end{groupplot}
\end{tikzpicture}
\caption{Attention entropy and the presence of a a strong attention sink both cluster by presence of QK norm. \poolname models with sliding window attention have less attention sink behavior on full attention layers; these represent the lower cluster of blue dots.} %
\label{fig:entropy_and_sinks}

\end{figure*}
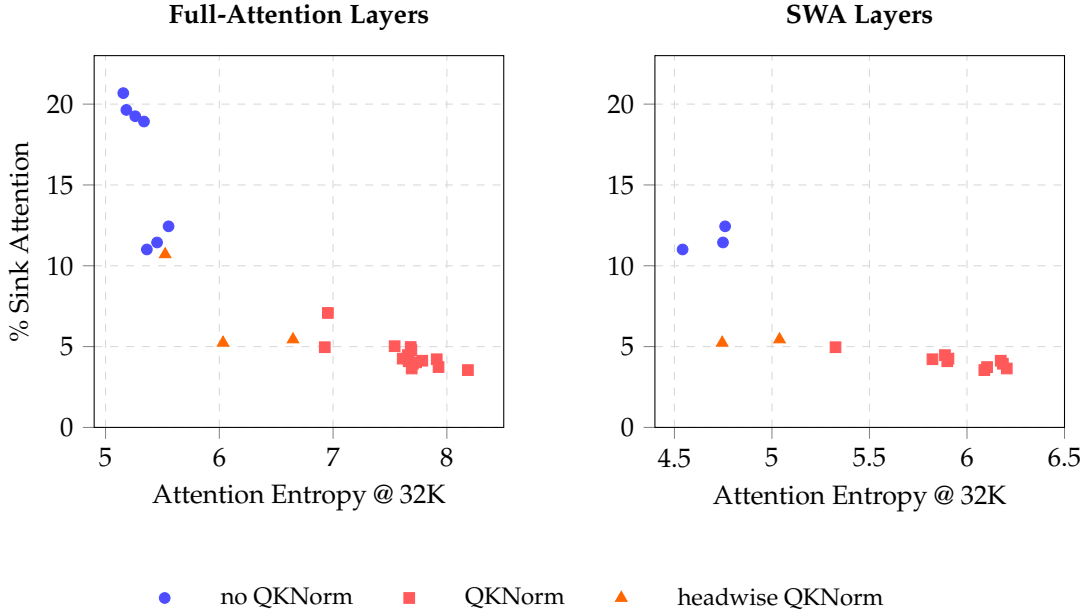
\paragraph{High entropy and attention sinks are positive indicators --- and QK norm reduces both.} \citet{yang2025rope} observe that models with QK-norm have higher entropy in the attention distribution (i.e. less peaky attention), which they suggest makes it more challenging for the model to attend over longer contexts. %
Attention entropy is also influenced by the presence of attention sinks \citep{xiao2024efficientstreaminglanguagemodels}: positions early in the context window that consistently receive substantial attention mass. The presence of these sinks is theorized to be a result of models attempting to ``discard'' excess attention weight \citep{bondarenko2023quantizabletransformersremovingoutliers, qiu2026unifiedviewattentionresidual, gu2025attentionsinkemergeslanguage} and avoid representation collapse \citep{barbero2025llmsattendtoken}; it is also believed to make models more difficult to quantize \citep{ye2025differentialtransformer}.

Figure~\ref{fig:entropy_and_sinks} visualizes both sink attention and attention entropy. While the presence of attention sinks is generally considered negative (e.g. \citet{qiu2025gated} names reduction of attention sinks as a core benefit of their approach), the presence of attention sinks here correlates with improved long context performance ($R^2=0.38$).
In the absence of another mechanism such as differential attention \citep{ye2025differentialtransformer} or gating \citep{qiu2025gated}, attention sinks appear to be the default strategy learned by QK-norm-less transformers to compensate for excess attention. Thus, attention sink behavior corresponds with long context abilities in \poolname. %

\paragraph{Retrieval heads} \citet{wu2024retrievalheadmechanisticallyexplains} propose the existence of \textit{retrieval heads}, specialized attention heads that are primarily responsible for retrieving information from prior context in both short and long context processing. %
We measure this using the same set of documents with a needle-in-a-haystack (NIAH) task injected.  We compute, at the end of prefill, the percentage of attention that is placed on the needle tokens. Then, we generate a completion for each example and look at examples where the model successfully generates the needle text. We compute per-head retrieval scores by measuring how much each individual attention head attended to the needle tokens in the input over the course of generation.

At prefill time, we observe no more than a slight correlation between the ability to place more attention on the needle tokens and performance on downstream evaluations. Models with QK norm (headwise \textit{or} layerwise) uniformly place less attention on the needle tokens than models without QK norm. However, during generation we observe little difference in retrieval head behavior across models in \poolname, with very low retrieval head scores across models. %
These models may be too weak to reliably identify retrieval heads. Alternatively, some other mechanism may govern long context abilities in models early in training. %

\section{Related work}
\paragraph{Architectural choices and long-context extensibility.} Prior works in this area primarily focus on positional embeddings or modifications to the extension method itself. The closest work to our setting, \citet{yang2025rope}, compares RoPE, NoPE, and QK-normalized RoPE models and shows that these three variants can differ substantially on long-context evaluation despite similar standard performance.
More broadly, work on positional design and extrapolation shows that even a single architectural axis can have a large effect on longer-range behavior: \citet{kazemnejad2023impact}, \citet{press2022train}, and \citet{wang2024length} find substantial differences across positional embedding variants, and \citet{lu2024controlled} find that sparse attention generally lags full attention for long context. %
Other work studies how to improve extrapolation by changing attention or positional embedding settings during context extension \citep{sun2023length, chen2023extending}. A separate line of long-context work changes the architecture, for example by modifying the attention mechanism \citep{zimerman2023long}, through recurrence and memory \citep{dai2019transformer}, or newer architectures designed for effectively unbounded context \citep{ma2024megalodon}; see \citet{huang2023advancing} for a survey of this space. In contrast, we focus on choices made before pretraining that may \textit{unintentionally} determine downstream long-context extension outcomes.

    \paragraph{Performance prediction from smaller scales.} A separate methodological literature asks how to make expensive model-development decisions using smaller or earlier experiments. Controlled pretraining suites such as Pythia \citep{biderman2023pythia}, open-science efforts such as OLMo and LLM360 \citep{groeneveld2024olmo,liu2023llm360}, and small-experiment prediction work such as DataDecide \citep{magnusson2025datadecide} show that many choices can be studied scientifically before full-scale deployment; scaling-law work shows that even expensive allocation decisions can be forecast from smaller runs \citep{hoffmann2022training}. %
    Some recent work has also looked at integrating features of the data distribution or architecture into scaling law predictions for short-context tasks \citep{liu2026notjustscalinglawsbetterunderstanding}.  \citet{magnusson2024paloma,fang2025wrong} argue that easy scalar proxies such as perplexity can miss important behavior during training runs,  while recent work on evaluation reliability shows that the signal-to-noise properties of benchmarks themselves can strongly affect how useful small experiments are for model-development decisions \citep{heineman2025signal}. More general work on capability prediction emphasizes that downstream behavior is often harder to forecast from pretraining signals alone \citep{schaeffer2025predicting}. Our setting is far cheaper than full-scale long context training, but more expensive than measuring pretraining signals alone. We also hope that \poolname can serve as a set of models on which to evaluate future performance prediction metrics. %

\section{Conclusion}
We demonstrate that a series of small, individually reasonable architectural perturbations, well grounded in the literature, can result in dramatically reduced long context capabilities.
We show that this degradation is difficult to detect in short context metrics but detectable from context extension runs very early into pretraining. This suggests two interesting directions for future research: evaluating the minimal size or token budget at which these effects can be reliably measured, to further reduce the cost of architectural experimentation; and devising better proxy metrics for short context models to estimate long context performance without performing a context extension. We also believe there is more to understand mechanistically about the differences between \poolname models. Finally, \poolname's parallel runs, traversing a similar optimization problem with slightly different architectures, may be useful for research into other phenomena in early pretraining. %
To this end, we release 38 %
checkpoints for each model, representing the full pretraining and long context extension. 

 Each feature we ablate has some clear benefit--- stability for normalization, pretraining efficiency for context length, and inference efficiency for sliding window and GQA--- and individually may be justified because of these other factors. Yet the combination of these features results in unacceptable long context extensibility. By exposing the interplay between these factors in a controlled setting, we hope to enable model developers to make more informed choices about their architecture design and to spur future research into alternatives that better navigate these tradeoffs. %

\section*{Ethics Statement}

Pretraining ablations carry a heavy computational cost.
We estimate that the cost of training the \poolsize models for this paper was approximately 170,000 H100 hours, in addition to the costs of initial experimentation, evaluation, and additional ablation runs. Using the calculations from \citet{Morrison2025HolisticallyET}, this is equivalent to approximately 42.63 metric tons of CO2 emitted, if our hardware was of equivalent power usage.  
In releasing all artifacts including intermediate checkpoints, we hope that this cost can be amortized by the reuse of \poolname.

\section*{Acknowledgements}

We thank Sewon Min, Adithya Pratapa, Prasann Singhal, Jacob Morrison, and the anonymous reviewers for helpful feedback on this work, and Taira Anderson, Kyle Wiggers, and Bailey Kuehl for release assistance.

This material is based upon work supported by the National Science Foundation under Award No. 2413244. AB was supported by a grant from the National Science Foundation Graduate Research Fellowship Program
under Grant No. DGE2140739. Any opinions, findings, and conclusions or recommendations expressed in this material are those of the author(s) and do not necessarily reflect the views of the sponsors.

\bibliography{colm2026_conference}
\bibliographystyle{colm2026_conference}

\newpage
\appendix
\section{\poolname}
\label{appendix:pool_desc}
Tables~\ref{tab:all_runs_helmet} and ~\ref{tab:all_runs_ruler} describe all \poolsize models in \poolname. The architectural columns of the table are identical, and replicated in the second table solely for ease of reading; Table~\ref{tab:all_runs_helmet} reports HELMET and LongPPL and Table~\ref{tab:all_runs_ruler} reports RULER scores for all runs. Both tables are sorted in order of HELMET score at 32K, with the best score bolded and the worst score italicized in each column. 

The column labeled SWA indicates training with sliding window layers: three out of every four layers use a 4096 local window in lieu of full attention. The QKNorm column indicates the presence of layerwise ($\checkmark$) or headwise (\texttt{hw}) QK norm. The initializations for each run is labeled with a letter code. The number of KV heads indicates the presence and degree of GQA used: 32 KV heads indicates full MHA. The feedforward dimension $d_{ff}$ was adjusted for some runs with GQA to keep the parameter counts as close as possible across training runs. 
\newpage
\begin{table*}[t]
\centering
\small
\setlength{\tabcolsep}{4pt}
\begin{tabular}{c c c c c r r r r | r r r r | r}
\toprule
\multicolumn{9}{c|}{\bf{Architecture}} & \multicolumn{4}{c|}{\bf{HELMET}} & \multirow{2}{*}{\bf{LongPPL}} \\
SWA & QKNorm & Norm & fp8 & Init & KV & Ctx & $d_\mathrm{ff}$ & Params & 8K & 16K & 32K & 64K & \\
\midrule
$\checkmark$ & hw & post & $\checkmark$ & A & 8 & 8K & 13,312 & 7.4B & \textit{42.9} & \textit{34.0} & \textit{29.9} & \textit{27.0} & 4.94 \\
$\checkmark$ & $\checkmark$ & post & \texttimes & B & 32 & 4K & 11,008 & 7.3B & 48.1 & 41.9 & 35.0 & 30.1 & 5.08 \\
$\checkmark$ & $\checkmark$ & post & \texttimes & C & 8 & 8K & 13,312 & 7.4B & 45.1 & 40.1 & 35.2 & 32.1 & 5.05 \\
$\checkmark$ & $\checkmark$ & post & \texttimes & H & 8 & 4K & 13,312 & 7.4B & 50.0 & 44.1 & 36.9 & 31.8 & 5.16 \\
$\checkmark$ & $\checkmark$ & post & \texttimes & D & 8 & 8K & 13,312 & 7.4B & 49.5 & 44.8 & 38.8 & 33.2 & 4.83 \\
$\checkmark$ & $\checkmark$ & post & \texttimes & G & 8 & 4K & 14,336 & 7.8B & 49.0 & 43.0 & 39.9 & 34.4 & \textit{5.21} \\
$\checkmark$ & $\checkmark$ & post & $\checkmark$ & D & 8 & 8K & 13,312 & 7.4B & 56.4 & 50.2 & 42.7 & 36.1 & 4.74 \\
$\checkmark$ & $\checkmark$ & post & $\checkmark$ & E & 32 & 8K & 11,008 & 7.3B & 54.6 & 48.7 & 42.8 & 37.4 & 4.31 \\
\texttimes & $\checkmark$ & post & \texttimes & B & 32 & 4K & 11,008 & 7.3B & 53.8 & 50.9 & 42.9 & 32.8 & 3.83 \\
$\checkmark$ & \texttimes & pre & \texttimes & F & 8 & 8K & 14,336 & 7.8B & 53.7 & 48.2 & 43.6 & 39.2 & 4.49 \\
$\checkmark$ & \texttimes & pre & \texttimes & G & 8 & 8K & 14,336 & 7.8B & 54.1 & 49.1 & 44.8 & 36.1 & 4.66 \\
$\checkmark$ & $\checkmark$ & pre & \texttimes & H & 32 & 8K & 11,008 & 7.3B & 57.2 & 51.5 & 46.3 & 39.1 & 4.67 \\
$\checkmark$ & $\checkmark$ & post & $\checkmark$ & H & 32 & 8K & 11,008 & 7.3B & 55.9 & 52.8 & 46.7 & 39.6 & 4.23 \\
\texttimes & hw & pre & \texttimes & K & 8 & 8K & 12,288 & 7.0B & 56.9 & 54.1 & 46.9 & 40.5 & 3.47 \\
$\checkmark$ & hw & post & \texttimes & H & 32 & 8K & 11,008 & 7.3B & 55.0 & 48.9 & 46.9 & 43.4 & 4.62 \\
\texttimes & $\checkmark$ & post & \texttimes & H & 32 & 4K & 11,008 & 7.3B & 54.3 & 52.3 & 47.4 & 37.6 & 3.76 \\
$\checkmark$ & $\checkmark$ & post & \texttimes & H & 32 & 8K & 11,008 & 7.3B & 58.4 & 52.5 & 47.5 & 41.1 & 4.38 \\
\texttimes & $\checkmark$ & post & \texttimes & G & 8 & 8K & 14,336 & 7.8B & 58.2 & 53.0 & 48.5 & 39.6 & 3.85 \\
\texttimes & $\checkmark$ & post & \texttimes & H & 32 & 8K & 11,008 & 7.3B & 57.7 & 54.3 & 48.7 & 39.7 & 3.70 \\
\texttimes & \texttimes & pre & \texttimes & G & 4 & 8K & 14,336 & 7.7B & 55.5 & 53.8 & 49.1 & 43.5 & 3.63 \\
\texttimes & \texttimes & pre & \texttimes & G & 8 & 4K & 14,336 & 7.8B & 58.5 & 55.2 & 50.2 & 42.1 & 3.78 \\
\texttimes & $\checkmark$ & pre & \texttimes & G & 8 & 8K & 14,336 & 7.8B & 59.4 & 55.5 & 51.4 & 42.4 & 3.70 \\
\texttimes & \texttimes & pre & \texttimes & G & 8 & 8K & 14,336 & 7.8B & 57.8 & 54.5 & 52.4 & 48.4 & 3.57 \\
$\checkmark$ & \texttimes & pre & \texttimes & H & 32 & 8K & 11,008 & 7.3B & 59.6 & 57.4 & 53.7 & 47.8 & 4.28 \\
\texttimes & \texttimes & pre & \texttimes & I & 32 & 8K & 12,288 & 7.8B & 59.3 & 55.5 & 54.4 & 49.3 & \bf{3.41} \\
\texttimes & \texttimes & pre & \texttimes & J & 16 & 8K & 14,336 & 8.1B & \bf{60.8} & \bf{58.6} & \bf{56.4} & \bf{50.0} & 3.50 \\
\bottomrule
\end{tabular}
\caption{All runs sorted by HELMET 32K (worst to best): HELMET scores and LongPPL.}
\label{tab:all_runs_helmet}
\end{table*}

\begin{table*}[t]
\centering
\small
\setlength{\tabcolsep}{4pt}
\begin{tabular}{c c c c c r r r r | r r r r r}
\toprule
\multicolumn{9}{c|}{\bf{Architecture}} & \multicolumn{5}{c}{\bf{RULER}} \\
SWA & QKNorm & Norm & fp8 & Init & KV & Ctx & $d_\mathrm{ff}$ & Params & 4K & 8K & 16K & 32K & 64K \\
\midrule
$\checkmark$ & hw & post & $\checkmark$ & A & 8 & 8K & 13,312 & 7.4B & 80.5 & 68.3 & 55.4 & 45.7 & 38.8 \\
$\checkmark$ & $\checkmark$ & post & \texttimes & B & 32 & 4K & 11,008 & 7.3B & 81.1 & 67.0 & 54.7 & 45.7 & 35.2 \\
$\checkmark$ & $\checkmark$ & post & \texttimes & C & 8 & 8K & 13,312 & 7.4B & 76.3 & \textit{60.5} & \textit{51.9} & 45.0 & 39.6 \\
$\checkmark$ & $\checkmark$ & post & \texttimes & H & 8 & 4K & 13,312 & 7.4B & 78.2 & 65.3 & 55.6 & \textit{44.4} & \textit{35.0} \\
$\checkmark$ & $\checkmark$ & post & \texttimes & D & 8 & 8K & 13,312 & 7.4B & 79.4 & 64.2 & 57.0 & 46.8 & 38.2 \\
$\checkmark$ & $\checkmark$ & post & \texttimes & G & 8 & 4K & 14,336 & 7.8B & \textit{76.2} & 63.6 & 54.9 & 48.8 & 40.2 \\
$\checkmark$ & $\checkmark$ & post & $\checkmark$ & D & 8 & 8K & 13,312 & 7.4B & 80.3 & 70.7 & 61.1 & 52.2 & 45.6 \\
$\checkmark$ & $\checkmark$ & post & $\checkmark$ & E & 32 & 8K & 11,008 & 7.3B & 81.0 & 71.0 & 62.8 & 54.6 & 45.8 \\
\texttimes & $\checkmark$ & post & \texttimes & B & 32 & 4K & 11,008 & 7.3B & 83.5 & 76.6 & 69.0 & 57.0 & 40.9 \\
$\checkmark$ & \texttimes & pre & \texttimes & F & 8 & 8K & 14,336 & 7.8B & 81.6 & 70.6 & 63.9 & 55.3 & 49.2 \\
$\checkmark$ & \texttimes & pre & \texttimes & G & 8 & 8K & 14,336 & 7.8B & 83.2 & 73.7 & 64.9 & 55.1 & 47.1 \\
$\checkmark$ & $\checkmark$ & pre & \texttimes & H & 32 & 8K & 11,008 & 7.3B & 84.6 & 74.3 & 64.1 & 55.8 & 49.5 \\
$\checkmark$ & $\checkmark$ & post & $\checkmark$ & H & 32 & 8K & 11,008 & 7.3B & 79.5 & 71.8 & 64.2 & 56.7 & 48.9 \\
\texttimes & hw & pre & \texttimes & K & 8 & 8K & 12,288 & 7.0B & 83.9 & 76.7 & 69.4 & 62.9 & 52.9 \\
$\checkmark$ & hw & post & \texttimes & H & 32 & 8K & 11,008 & 7.3B & 81.1 & 68.9 & 61.8 & 57.4 & 55.8 \\
\texttimes & $\checkmark$ & post & \texttimes & H & 32 & 4K & 11,008 & 7.3B & 81.5 & 74.3 & 66.7 & 60.9 & 49.5 \\
$\checkmark$ & $\checkmark$ & post & \texttimes & H & 32 & 8K & 11,008 & 7.3B & 81.2 & 70.4 & 61.2 & 54.1 & 48.4 \\
\texttimes & $\checkmark$ & post & \texttimes & G & 8 & 8K & 14,336 & 7.8B & 79.8 & 74.6 & 64.6 & 57.1 & 44.2 \\
\texttimes & $\checkmark$ & post & \texttimes & H & 32 & 8K & 11,008 & 7.3B & 84.9 & 75.0 & 66.8 & 59.3 & 50.8 \\
\texttimes & \texttimes & pre & \texttimes & G & 4 & 8K & 14,336 & 7.7B & 80.7 & 73.3 & 67.5 & 61.2 & 54.0 \\
\texttimes & \texttimes & pre & \texttimes & G & 8 & 4K & 14,336 & 7.8B & 81.1 & 67.0 & 54.7 & 45.7 & 35.2 \\
\texttimes & $\checkmark$ & pre & \texttimes & G & 8 & 8K & 14,336 & 7.8B & 82.5 & 75.6 & 68.3 & 62.8 & 50.7 \\
\texttimes & \texttimes & pre & \texttimes & G & 8 & 8K & 14,336 & 7.8B & 83.2 & 78.1 & 72.0 & 64.6 & 56.0 \\
$\checkmark$ & \texttimes & pre & \texttimes & H & 32 & 8K & 11,008 & 7.3B & 82.8 & 74.6 & 66.2 & 64.2 & 57.3 \\
\texttimes & \texttimes & pre & \texttimes & I & 32 & 8K & 12,288 & 7.8B & 85.2 & \bf{81.6} & 75.5 & 67.5 & 58.9 \\
\texttimes & \texttimes & pre & \texttimes & J & 16 & 8K & 14,336 & 8.1B & \bf{86.0} & 79.9 & \bf{76.0} & \bf{67.7} & \bf{64.9} \\
\bottomrule
\end{tabular}
\caption{All runs sorted by HELMET 32K (worst to best): RULER scores. }
\label{tab:all_runs_ruler}
\end{table*}

\clearpage
\section{Additional features studied}
\label{appendix:additional_features}

\paragraph{Norm ordering and type of QK norm have small effects.} 

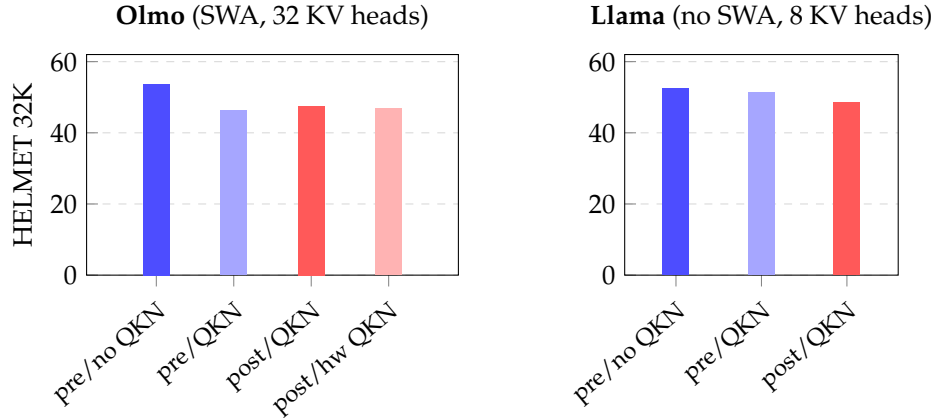
\begin{figure*}[t]
\centering
\begin{tikzpicture}
\begin{groupplot}[
    group style={
        group size=2 by 1,
        horizontal sep=2.2cm,
    },
    ymin=0, ymax=62,
    ybar,
    ylabel={HELMET 32K},
    xtick align=outside,
    tick pos=left,
    ymajorgrids=true,
    xmajorgrids=false,
    grid style={dashed, gray!30},
    enlarge x limits=0.3,
    x tick label style={rotate=40, anchor=north east, font=\footnotesize},
]

\nextgroupplot[
    title={\textbf{Olmo} (SWA, 32 KV heads)},
    width=6.5cm,
    height=4.5cm,
    symbolic x coords={A, B, C, D},
    xtick={A, B, C, D},
    xticklabels={pre/no QKN, pre/QKN, post/QKN, post/hw QKN},
]
\addplot[fill=blue!70, draw=none, bar shift=0pt] coordinates {(A, 53.69)};
\addplot[fill=blue!35, draw=none, bar shift=0pt] coordinates {(B, 46.32)};
\addplot[fill=red!65,  draw=none, bar shift=0pt] coordinates {(C, 47.55)};
\addplot[fill=red!30,  draw=none, bar shift=0pt] coordinates {(D, 46.88)};

\nextgroupplot[
    title={\textbf{Llama} (no SWA, 8 KV heads)},
    ylabel={},
    width=5.2cm,
    height=4.5cm,
    symbolic x coords={A, B, C},
    xtick={A, B, C},
    xticklabels={pre/no QKN, pre/QKN, post/QKN},
]
\addplot[fill=blue!70, draw=none, bar shift=0pt] coordinates {(A, 52.38)};
\addplot[fill=blue!35, draw=none, bar shift=0pt] coordinates {(B, 51.37)};
\addplot[fill=red!65,  draw=none, bar shift=0pt] coordinates {(C, 48.55)};

\end{groupplot}
\end{tikzpicture}
\caption{QK norm is harmful for long context (as first observed by \citet{yang2025rope}); we further note that the headwise variant is an additional slight detriment to downstream performance, and the less harmful norm order with QK norm is model-family-dependent.}
\label{fig:norms}
\end{figure*}

We ablate the three combinations of QK norm and norm order that are stable in our training regime: preorder without QK norm, preorder with QK norm, and post-sublayer order with QK norm in Figure~\ref{fig:norms}.
We find that norm order alone has an inconsistent effect, with QK norm accounting for the vast majority of the performance difference between runs. Headwise QK norm causes an additional slight degradation. QK norm is most often added to improve training stability (e.g. \citet{chameleonteam2025chameleonmixedmodalearlyfusionfoundation, marin2025retro32b}); we also confirm that it improves stability in our setting in Appendix~\ref{appendix:stability}.

\paragraph{Float8 pretraining does not appear to have a consistent effect.} We consider training linear layers in float8 instead of bfloat16 precision. In two controlled comparisons, adding float8 training results in a slight degradation in one comparison and an improvement in the other. We find no evidence that float8 pretraining degrades long context performance--- instead, it appears that float8's effect in \poolname models may be purely noise, like initialization, because float8 optimization results in a slightly different optimization path. We note that we do not consider a setting where attention weights are quantized during pretraining; it is possible that this would result in degradation, as long context abilities inherently require the model to be able to attend precisely over a long context window.

 \paragraph{Initialization causes more variation in short context than long context.} We are not able to completely standardize initialization across runs because some models differ in parameterization. In \poolname, we construct four pairs of runs that are identical except for initialization and measure the swing in performance for both short context and long context metrics. To standardize across metrics with differing scales, we compute the swing due to initialization as a percentage of the maximum variation between runs in \poolname. 
 
 Initialization has minimal impact on train loss and perplexity (shifting runs by less than 3\% of the cross-run range on average) and causes the largest swings in the short context benchmarks and pre-context-extension HELMET scores, where runs with the same architecture but different initialization may vary by up to 58\% of the total range of variance. 
 This effect is reduced after long context extension; on the long context benchmarks, initialization can account for smaller but still substantial swings of up to 17\% (and on average 7.7\%) of the observed range of values. In the results, we discuss only differences across architectural variations that result, on average, in long context score shifts substantially larger than the mean shift from  initialization. Note that the sole dimension where we cannot construct an initalization-controlled trial is in measuring the impact of varying GQA degree; thus, the effect size for the difference there may be slightly larger or smaller than the one we observe. %

\section{Additional Training Details}
\label{appendix:training_details}

For additional details on training and the pretraining data used, we refer the reader to the Olmo 3 technical report \citep{olmo3_2025}; we follow the configuration for Olmo 3 7B Base pretraining, although we use substantially less than 1024 concurrent GPUs for these much shorter runs. The \href{https://github.com/allenai/olmpool/}{project github repository}
provides configurations for each pretraining and context extension run, and is the best reference point for specific training details for each run.

\paragraph{On optimization hyperparameters.}
We use the same learning rate, batch size, and optimization schedule for all models in \poolname.
While it would be infeasible to sweep all hyperparameters for all models in \poolname, we believe that our selections for these hyperparameters are reasonable for all models in the pool for several reasons. First, because we only pretrain for 140B tokens in a 5T learning rate schedule, changing the learning rate schedule causes only marginal differences. For instance, we trained one development run at both 5T and 7T learning rate schedules, and observed near-identical performance on both loss and downstream metrics.
We do not change the depth of networks\footnote{With one exception: the Qwenlike model is 36 instead of 32 layers to match Qwen} and hold parameter count as close to constant as possible, standardizing two major factors often identified as driving changes in optimal learning rate \citep{hyperparam_transfer}.

Additionally, several works observe that batch size has minimal impact below a critical batch size at which learning slows \citep{mccandlish2018empiricalmodellargebatchtraining}. We are below the estimated critical batch size for models of this configuration \citep{merrill2025criticalbatchsizerevisited} for the vast majority of training, and we have no reason to believe that any of these changes would impact the critical batch size; in fact, \citet{zhang2025doescriticalbatchsize} show empirically that context length does not impact critical batch size.

\paragraph{Skip-step optimizer}
A final important detail that we highlight here: like all recent Olmo models, \poolname models are trained with a skip-step optimizer, which skips steps that have an abnormally high gradient norm to increase run stability. This triggers very rarely, but when it does occur, it causes a small amount of variance in pretraining data across \poolname (i.e., for a small percentage of data, not every model performs a gradient update). For more details, see the
\href{
https://olmo-core.readthedocs.io/en/latest/optim.html#olmo_core.optim.SkipStepOptimizer}{documentation}.

\section{Further evaluation details}
\label{appendix:evaluation_details}
This appendix briefly describes each short-context evaluation metric and validation perplexity set and its correlation with long context downstream.

\subsection{Loss}

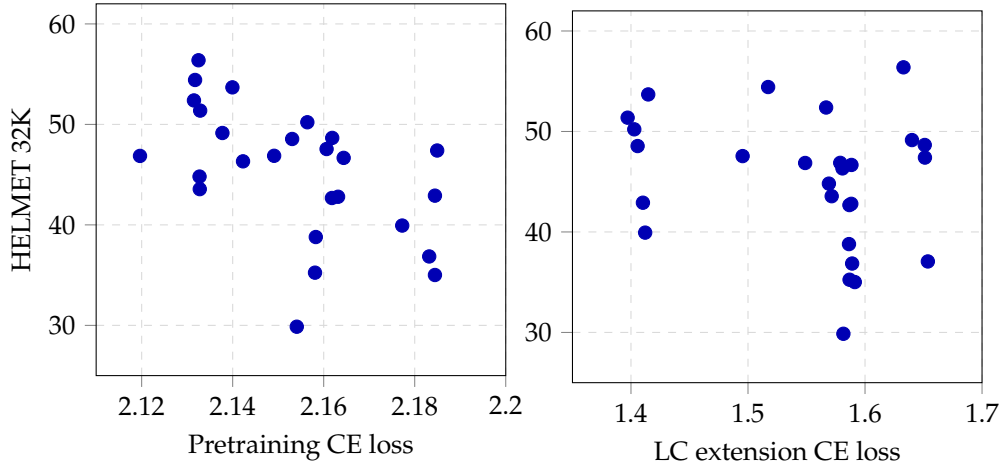
\begin{figure*}[h!]
    \centering
    \begin{subfigure}[t]{0.48\textwidth}
            \begin{tikzpicture}
\begin{axis}[
    width=7cm,
    height=6.5cm,
    xlabel={Pretraining CE loss},
    ylabel={HELMET 32K},
    xmin=2.11, xmax=2.20,
    ymin=25,   ymax=62,
    tick align=outside,
    tick pos=left,
    grid=major,
    grid style={dashed, gray!30},
]

\addplot[only marks, mark=*, mark size=2.5pt, color=blue!70!black]
    table[col sep=comma, x=train_loss, y=HELMET_32K]
    {figures/data/train_loss_vs_helmet.csv};

\end{axis}
\end{tikzpicture}  %

    \end{subfigure}
    \begin{subfigure}[t]{0.48\textwidth}
            \begin{tikzpicture}
\begin{axis}[
    width=7cm,
    height=6.5cm,
    xlabel={LC extension CE loss},
    xmin=1.35, xmax=1.70,
    ymin=25,   ymax=62,
    tick align=outside,
    tick pos=left,
    grid=major,
    grid style={dashed, gray!30},
]

\addplot[only marks, mark=*, mark size=2.5pt, color=blue!70!black]
    table[col sep=comma, x=lc_loss, y=HELMET_32K]
    {figures/data/lc_loss_vs_helmet.csv};

\end{axis}
\end{tikzpicture}
            \label{subfig:helmet_sc}
            
    \end{subfigure}
   \caption{Training loss does not correlate well with long context ability, in either the pretraining or long context extension runs.}
    \label{fig:loss_vs_lc}

\end{figure*}

In Figure~\ref{fig:loss_vs_lc}, we graph the (slight) correlation between loss and long context ability. Both pretraining and long context loss fail to correlate with long context capabilities.

\subsection{Perplexity on held-out text}

 We measure perplexity across a diverse set of held-out validation splits, drawn mostly from Paloma \citep{magnusson2024paloma}.
  \textbf{C4} \citep{raffel2020c4} is a filtered and deduplicated web corpus
  derived from Common Crawl; we use the English validation split.
  \textbf{Dolma} \citep{soldaini-etal-2024-dolma} is Ai2's open pretraining corpus;
  we evaluate on six domain-specific splits: books, Common Crawl, peS2o
  (scientific papers), Reddit, Stack Exchange, and Wikipedia.
  \textbf{ICE} \citep{greenbaum1996ice} (International Corpus of English) provides
  text spanning multiple varieties of dialectal English.
  \textbf{M2D2} \citep{reid2022m2d2} is a massively multi-domain dataset; we use
  the S2ORC split covering scientific text.
  \textbf{The Pile} \citep{gao2020pile} is an 800GB diverse text corpus from
  EleutherAI, evaluated on its held-out validation set.
  \textbf{WikiText-103} \citep{merity2016pointer} is a standard language modeling
  benchmark of Wikipedia articles.

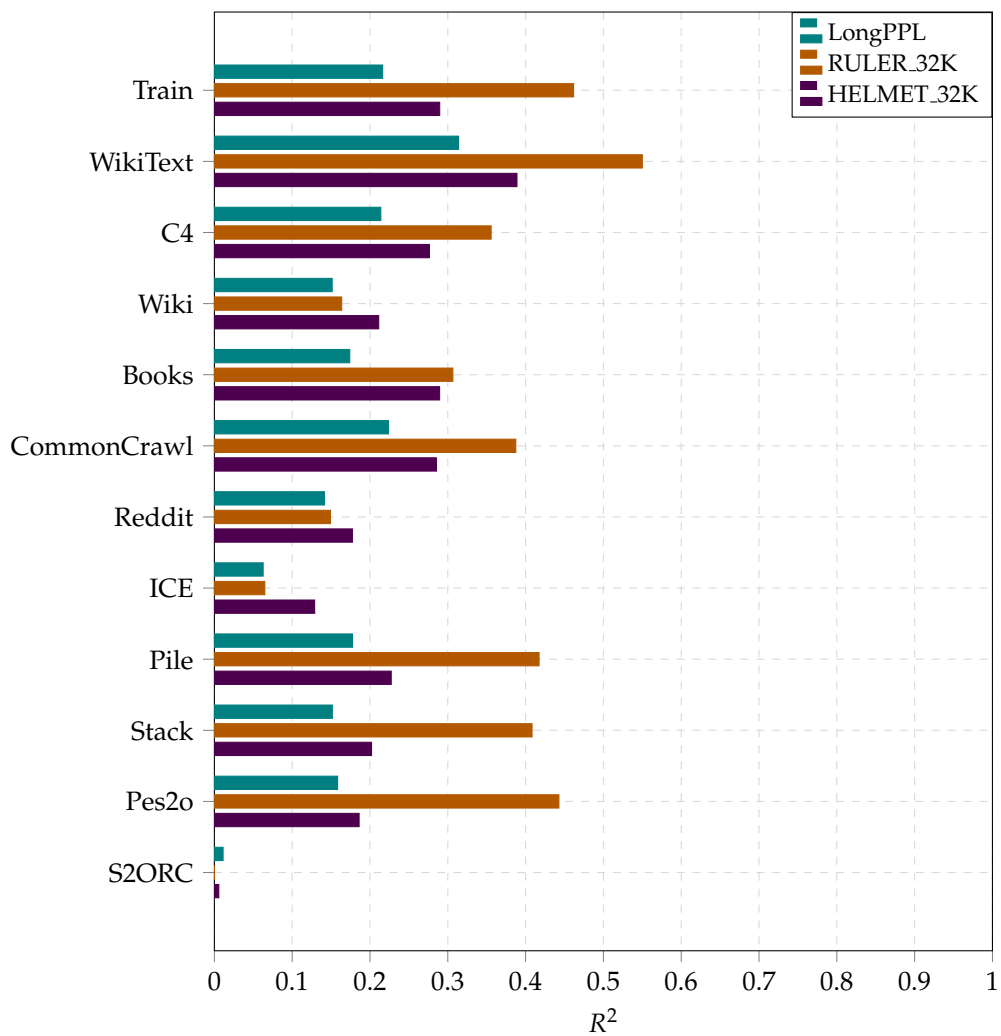
\begin{figure*}[t]
\centering
\begin{tikzpicture}
\begin{axis}[
  xbar,
  width=0.85\textwidth,
  height=14cm,
  bar width=5pt,
  xmin=0, xmax=1,
  xlabel={$R^2$},
  ytick=data,
  yticklabels={Train,WikiText,C4,Wiki,Books,CommonCrawl,Reddit,ICE,Pile,Stack,Pes2o,S2ORC},
  tick align=outside,
  tick pos=left,
  grid=major,
  grid style={dashed, gray!30},
  legend style={at={(1.0,1.0)}, anchor=north east, font=\small},
  legend cell align=left,
  legend entries={HELMET\_32K, RULER\_32K, LongPPL},
  legend reversed=true,
  y dir=reverse,
]

\addplot[fill=violet!60!black, draw=violet!60!black] coordinates {(0.2896,0) (0.3889,1) (0.2764,2) (0.2112,3) (0.2895,4) (0.2854,5) (0.1774,6) (0.1289,7) (0.2273,8) (0.2020,9) (0.1860,10) (0.0057,11)};

\addplot[fill=orange!70!black, draw=orange!70!black] coordinates {(0.4617,0) (0.5500,1) (0.3557,2) (0.1636,3) (0.3065,4) (0.3874,5) (0.1492,6) (0.0648,7) (0.4173,8) (0.4083,9) (0.4428,10) (0.0001,11)};

\addplot[fill=teal, draw=teal] coordinates {(0.2163,0) (0.3138,1) (0.2139,2) (0.1514,3) (0.1740,4) (0.2237,5) (0.1415,6) (0.0627,7) (0.1775,8) (0.1516,9) (0.1584,10) (0.0112,11)};

\end{axis}
\end{tikzpicture}
\caption{$R^2$ between each standard validation perplexity metric and downstream
  long-context metrics (HELMET 32K, RULER 32K, LongPPL).}
\label{fig:ppl_correlations}
\end{figure*}

Figure~\ref{fig:ppl_correlations} shows the per-split correlation with the downstream long context metrics. %

\newpage

\subsection{BPB on downstream benchmarks}
\label{appendix:short_context_metrics}

We evaluate short-context capabilities across several standard benchmarks.                          
  \textbf{ARC} (Easy and Challenge splits) \citep{clark2018arc} is grade-school
  science question answering, with the Challenge split requiring more complex
  reasoning. \textbf{HellaSwag} \citep{zellers2019hellaswag} measures commonsense
  natural language inference via sentence completion. \textbf{MMLU} \citep{hendrycks2021mmlu}
  measures graduate-level knowledge across 57 subjects; we use the general split into humanities, social sciences,
  STEM, and other domains . \textbf{HumanEval} \citep{chen2021evaluatinglargelanguagemodels} requires Python code generation conditioned on docstrings.
  \textbf{MBPP} \citep{austin2021mbpp} similarly evaluates code generation on
  entry-level Python programming problems. \textbf{MINERVA} \citep{lewkowycz2022minerva}
  is a set of 500 math problems requiring
  multi-step numerical and symbolic computation. Finally, \textbf{Basic Skills} \citep{olmo3_2025}
  is a datatset covering six fundamental competencies: arithmetic, coding,
  common knowledge, logical reasoning, pattern recognition, and string operations.
  All downstream metrics are reported as bits-per-byte (BPB) on the gold answer string.

\begin{figure*}[t]
\centering
\begin{tikzpicture}
\begin{axis}[
  xbar,
  width=0.85\textwidth,
  height=16cm,
  bar width=5pt,
  xmin=0, xmax=1,
  xlabel={$R^2$},
  ytick=data,
  yticklabels={SC\_avg,ARC-C,ARC-E,HellaSwag,MMLU-Hum,MMLU-Other,MMLU-SocSci,MMLU-STEM,HumanEval,MBPP,MATH,BasicSkills,BS-Arith,BS-Code,BS-Logic,BS-Pattern,BS-String},
  tick align=outside,
  tick pos=left,
  grid=major,
  grid style={dashed, gray!30},
  legend style={at={(1.0,1.0)}, anchor=north east, font=\small},
  legend cell align=left,
  legend entries={HELMET\_32K, RULER\_32K, LongPPL},
  legend reversed=true,
  y dir=reverse,
]

\addplot[fill=violet!60!black, draw=violet!60!black] coordinates {(0.1735,0) (0.2236,1) (0.0059,2) (0.1198,3) (0.3727,4) (0.1388,5) (0.2943,6) (0.1564,7) (0.3061,8) (0.0593,9) (0.3665,10) (0.1911,11) (0.0002,12) (0.0672,13) (0.0021,14) (0.0081,15) (0.1696,16)};

\addplot[fill=orange!70!black, draw=orange!70!black] coordinates {(0.1912,0) (0.1945,1) (0.0010,2) (0.0949,3) (0.3755,4) (0.0963,5) (0.2615,6) (0.1765,7) (0.2606,8) (0.0310,9) (0.3846,10) (0.1731,11) (0.0118,12) (0.0773,13) (0.0005,14) (0.0025,15) (0.1935,16)};

\addplot[fill=teal, draw=teal] coordinates {(0.2192,0) (0.2581,1) (0.0455,2) (0.0741,3) (0.3766,4) (0.1055,5) (0.2556,6) (0.1248,7) (0.1856,8) (0.0425,9) (0.3190,10) (0.2002,11) (0.0381,12) (0.0452,13) (0.0045,14) (0.0053,15) (0.2509,16)};

\end{axis}
\end{tikzpicture}
\caption{$R^2$ between each short-context benchmark metric and downstream
  long-context metrics (HELMET 32K, RULER 32K, LongPPL).}
\label{fig:sc_correlations}
\end{figure*}
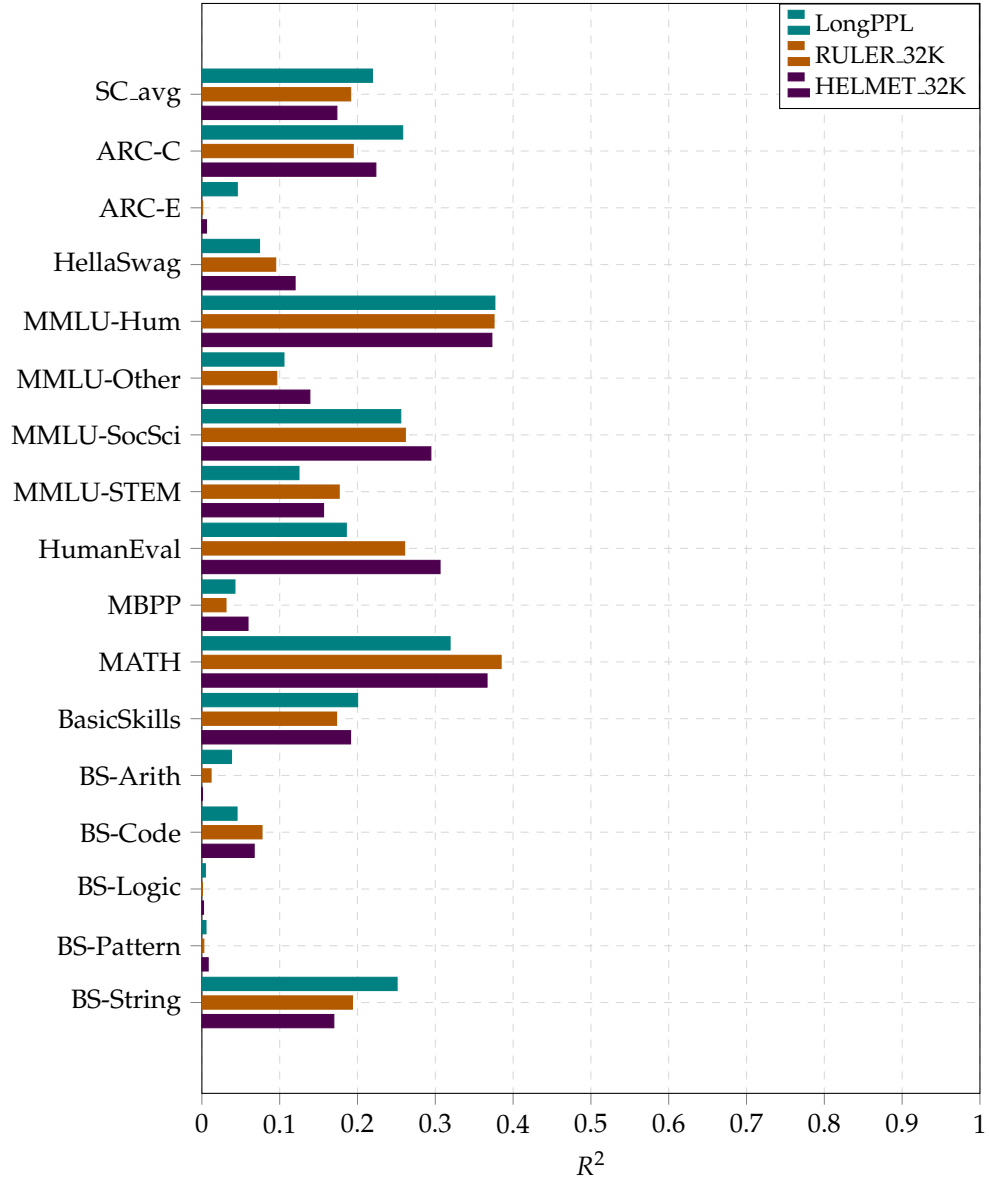

    Figure~\ref{fig:sc_correlations} shows the correlation between each individual short context benchmark BPB and the downstream long context scores. No benchmark correlates strongly, and we do not observe any consistent trend in which types of benchmarks correlate most. Note one of the three coding-related benchmarks correlates strongly, while the others do not. Since LongPPL correlates very poorly with validation perplexity for these datasets, we believe this is likely because these datasets contain a supermajority of tokens which can be predicted using only short-context dependencies; for more details on this problem, see 
    \citet{fang2025wrong}.

\clearpage
\subsection{Training stability} 
\label{appendix:stability}

Another important feature in pretraining is the stability of the training process. We measure a score for stability by computing the percentage of gradient norms that are more than 6 standard deviations from the mean during the pretraining run. We observe that runs with QK norm have, on average, a lower spike score, which aligns with prior observations that QK norm is beneficial for stability. 

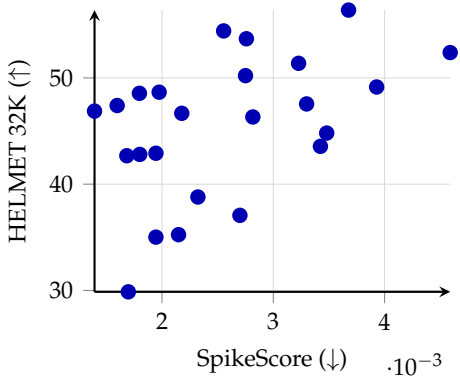
\begin{figure}[h]
    \centering
      \begin{tikzpicture}
    \begin{axis}[
      lc arch,
      xlabel={SpikeScore ($\downarrow$)},
      ylabel={HELMET 32K ($\uparrow$)},
    ]

      \addplot[only marks, mark=*, color=blue!70!black]
        table[col sep=comma, x=SpikeScore, y=HELMET_32K]
        {figures/data/plot_spike_helmet.csv};

    \end{axis}
  \end{tikzpicture}
  
    \caption{Spike score vs.\ HELMET 32K across \poolname. Stability weakly correlates with worse LC performance downstream, mostly due to QK-norm's influence on both factors.}
  \label{fig:spike_vs_helmet}
\end{figure}

We then calculate correlation between this score and long context performance downstream and visualize this in Figure~\ref{fig:spike_vs_helmet}. In \poolname, there is a slight negative correlation ($R^2=0.22)$ between pretraining stability and long context performance, mostly because QK norm both improves stability and damages long context performance.

\end{document}